\pdfoutput=1

\documentclass[11pt]{article}

\usepackage[review]{naacl2021}

\usepackage{times}
\usepackage{latexsym}
\usepackage{graphicx}
\usepackage{booktabs}

\usepackage[T1]{fontenc}

\usepackage[utf8]{inputenc}

\usepackage{microtype}

\title{SocialNLP Fake-EmoReact 2021 Challenge Overview: Predicting Fake Tweets from Their Replies and GIFs}

 \author{Chien-Kun Huang$^{1}$\thanks{* denotes equal contribution}, \ Yi-Ting Chang$^{2*}$, \ Lun-Wei Ku${^1}$, \ Cheng-Te Li${^3}$,\ Hong-Han Shuai$^{2}$\\
         Institute of Information Science, Academia Sinica${^1}$  \\  
         National Yang Ming Chiao Tung University${^2}$, 
         National Cheng Kung University${^3}$  \\
         \{kunhuang, lwku\}@iis.sinica.edu.tw \\
         \{joshchang0111.ee06, hhshuai\}@nycu.edu.tw \\
        \{chengte\}@mail.ncku.edu.tw
         }

\begin{document}
\maketitle
\begin{abstract}
This paper provides an overview of the Fake-EmoReact 2021 Challenge, held at the 9th SocialNLP Workshop, in conjunction with NAACL 2021. The challenge requires predicting the authenticity of tweets using reply context and augmented GIF categories from EmotionGIF dataset. We offer the Fake-EmoReact dataset with more than 453k as the experimental materials, where every tweet is labeled with authenticity. Twenty-four teams registered to participate in this challenge, and 5 submitted their results successfully in the evaluation phase. The best team achieves 93.9 on Fake-EmoReact 2021 dataset using F1 score. In addition, we show the definition of share task, data collection, and the teams' performance that joined this challenge and their approaches.
\end{abstract}

\section{Introduction}
Accessing data from social media has gained much attention in recent years due to the development of the internet and social platform (e.g., Facebook and Twitter). With the convenience of sharing information on online platforms, fake news could be spread rapidly from one to another, especially on social media. The request for fact checking through textual information raises significantly with the growth of users \cite{wang2017liar}. The misinformation has been used to influence people's acknowledge, such as politicians can draw attention away from issues against them by using social media \cite{lewandowsky2020using}. The situation is even worse due to the pandemic of Covid-19. Seeing the high impact of fake news on society, we eager to mitigate the effect of fake news by applying NLP techniques.

Two main dataset for fact checking: FEVER \cite{thorne2018fever} and LIAR \cite{wang2017liar}. The former dataset uses Wikipedia sentences for verification, and the latter evaluated rumors by editors. However, it is rare to find evidence written in an informal
language from social media services to verify fake news.

\begin{figure}[t!]
    \centering
    \includegraphics[width=0.5\textwidth]{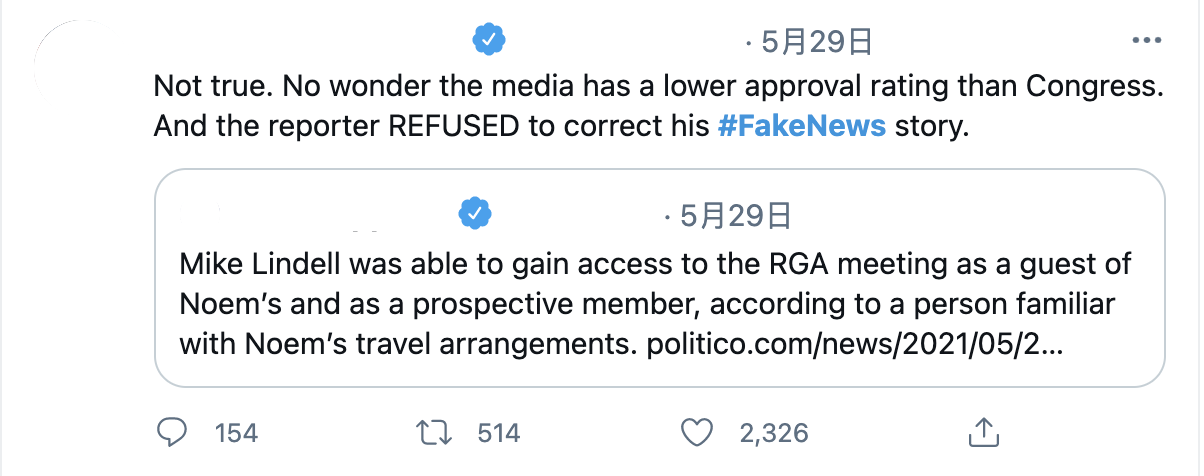}
    \caption{An example of fake new message on Twitter}
    \label{fig:intro}
\end{figure}

Most of the textual information available on social platforms is short and direct. Moreover, these messages also provide heterogeneous social information (e.g., tags, friends, followers, endorsements, profiles, and retweets) that should be considered together with the contents for better analysis. Figure \ref{fig:intro} shows an example of tweets message with fake news text, retweets, and other social information.

To improve research on fact checking and create a new research possibility, we proposed the Fake-EmoReact 2021 shared task. The competition provided a unique dataset of 453k tweets with replies and their fine-grained reaction category from EmotionGIF 2020 \cite{shmueli2021socialnlp}. The challenge is defined as given an unlabeled tweet and its GIF response, and the model should predict the label of a tweet as fake or real from the unlabeled evaluation dataset. More detail about Fake-EmoReact 2021 dataset, the competition, and the results are provided below.

\section{Fake-EmoReact dataset}

\begin{table*}[t]
\centering
\begin{tabular}{@{}llllll@{}}
\toprule
\textbf{idx} & \textbf{text} &\textbf{categories} &\textbf{context\_idx} &\textbf{reply} & \textbf{label} \\ 
\midrule

44025           & ``"It seems to me ...''          &['Facepalm', 'Oops',...]& 9 & ''@dustndiesel...'' & Fake          \\

44025           & ``"It seems to me ...          &[]& 10 & ''@AdamCoffeyNT...'' & Fake          \\

45600 & ``I got a ...''&['others'] & 14 &''@SteveHofstetter...'' & Real          \\
45600 & ``I got a ...'' &['Hearts', 'Hug', ...]& 15 & ''@SteveHofstetter...'' & Real          \\ 

\bottomrule
\end{tabular}
\caption{Samples of Fake-EmoReact 2021 Dataset}
\label{tab:data_sample}
\end{table*}

As social media arousing in recent years, the internet becomes an excellent platform for news dispersion. Twitter is one such platform that allows users to share any message despite its veracity directly. Nevertheless, this can cause a tremendous impact on the community since lots of them are either misinformation or unverified rumors. During the last two years, we have witnessed the severe situation brought by the pandemic of COVID-19. However, misleading information can even lead us to a much worse circumstance. Consequently, modeling to alleviate the impact of misinformation by automatically detecting fake news on social websites has attracted significant attention in the research field. For the reasons mentioned above, we create the Fake-EmoReact dataset, which consists of thousands of tweets from Twitter.

In the Fake-EmoReact dataset, tweets and corresponding conversational threads are collected, including all textual and animated GIF replies. It is worthy to note that all the tweets in this dataset contain at least one GIF response since we consider this type of posts can accurately convey its emotion towards the source tweet in an expressive way. For positive samples, We follow previously proposed Covid-19 fake news in tweets \cite{singh2020first}.
For source tweets with hashtag fake news ("\#FakeNews"), we remove the "\#FakeNews" and give the label as "fake". On the other hand, source tweets without "\#FakeNews" are labeled as "real". Each sample in the dataset contains the text of the original source tweet and the text of one of its reply tweets. We also provide the video files and their GIF categories if the reply contains a GIF. Given a GIF, we scan through the top-100 GIFs in each of the 43 pre-defined categories in Twitter to see which class it belongs to. For simplicity, if the GIF is not found in the top-100, we label it as "others". Note that some of the GIFs have multiple categories.
Several samples from Fake-EmoReact 2021 dataset was show in
Table \ref{tab:data_sample}.

In detail, the "idx" and "text" field in each sample represents the source tweets' index and text content, respectively. The "context\_idx" indicates the index of the response tweet to the source tweet, which is arranged according to the chronological order in the conversational thread, and "reply" is its text content. The field "categories" is the category of the GIF response, and "mp4" stands for the file name of the GIF video. And, "label" represents whether the source tweet is fake or not.

\textbf{Data Collection} The pipeline of our data collection can be roughly divided into three stages. In the beginning, we crawled the tweets whose text contain "\#FakeNews". Since the Twitter API doesn't provide a way to retrieve the replies under a specific tweet, we then crawled the reply tweet ids from Twitter's web interface. Lastly, we use the Twitter API to obtain the complete content of each reply and filter out the conversational threads which do not contain any GIF response. The collection of real samples follows the same procedure mentioned above, and the only difference is that the source tweets doesn't include "\#FakeNews". Specifically, samples with "real" labels in the training and development sets are combined from the EmotionGIF 2020 dataset \cite{shmueli2021socialnlp}. While in the evaluation set, we further collect the real samples in the same way as the fake samples.

\begin{table}[t]
\centering
\begin{tabular}{@{}ll|r|rrrrrrrr@{}}
\multicolumn{1}{c}{\textbf{}} & \multicolumn{1}{c|}{\textbf{}} & \multicolumn{1}{c|}{\textbf{\# text}} & \multicolumn{1}{c}{\textbf{\# reply}}\\ \midrule
\textbf{Fake-EmoReact}  & 
\textbf{Real}     & 43,200                                 &152,528                              \\
\textbf{}                     & \textbf{Fake}   & 7,200                                   &300,677                                    \\
\textbf{}                     & \textbf{Total}      & 50,400                                 & 453,205                                    \\
   
\end{tabular}
\caption{Label Distribution}
\label{tab:dataset}
\end{table}

\textbf{Label Distrubution}
Tables \ref{tab:dataset} shows the amount of sample in Fake-EmoReact 2021 dataset by label. More than 50k texts were provided, with real 43,200 and fake 7,200 respectively. In the dataset, one text may contain more than one reply. Therefore, the total number of replies is larger than the number of texts.

\section{Share Task}

\begin{table*}[t]
\centering
\begin{tabular}{@{}lllcrrrr@{}}
\toprule
\textbf{Rank} & \textbf{Team} & \textbf{Model} & \textbf{F1-macro} & \textbf{Precision} & \textbf{Recall}  \\ 
\midrule
1 & Yao & DeBERTa & 93.90 & 93.46 & 94.74\\
2 & Jina & LR, Textual Characteristics &  84.59 & 84.27 & 85.15 \\
3 & SpencerChen &      & 80.40 & 83.27 & 79.37 \\
4 & TeamZulu & Ensemble framework, Voting techniques& 79.81 & 79.93 & 79.71 \\
5 & skblaz &  & 71.40 & 80.62 & 70.77 \\
 
\bottomrule
\end{tabular}
\caption{F-scores for round two evaluation dataset (\%)}
\label{tab:model}

\end{table*}

We set up a specific website for Fake-EmoReact 2021 competition website\footnote{\url{https://sites.google.com/view/covidfake-emoreact-2021/home?authuser=0}}. The platform provides information about introduction, evaluation criteria, important date, registration form, and other competition details. Following registration, participants will be given a link to access Codalab\footnote{\url{https://codalab.org/}} to download datasets and upload their predictions.

In the share task, the dataset is split into training, development, and evaluation set with 168,522, 40,488, 88,665 tweets in round one, respectively. Additional development (45,037 tweets) and evaluation dataset (110,493 tweets) was given in round two. The label of tweets is removed from the development and evaluation dataset in both rounds. 
The goal of the share task is to predict the truthfulness of a tweet with its original context, reply text, and GIF categories. Macro-F1 score from scikit-learn\footnote{\url{https://scikit-learn.org/stable/}} was used as the competition metric in all rounds.

\section{Submission}
Twenty-four teams register in the shared task. Two groups participated successfully to the evaluation phase in Round One, which is optional for Fake-EmoReact 2021. And five teams successfully submitted entries in Round Two, with their results presenting on the Fake-EmoReact 2021 website. 
Of the top participants, five teams presented a technical report and shared their code, as was required by the competition rules. Pre-trained models was allowed in the share task. We use macro-F1 scores as our evaluation matrix by giving the label of tweet (Real and Fake) as class.

 In the following sections we briefly introduce the methods used by the teams, and conclude with a summary and some notes. Two of the submissions did not follow up with technical papers and thus they do not appear in this summary. More detail descriptions can been found in the teams’ technical reports.

\paragraph{Yao}
DeBERTa-base model \cite{he2020deberta} was used as a pre-trained model and fine-tuned to the Fake-EmoReact 2021 dataset with binary cross entropy loss. DeBERTa composes disentangled attention mechanism and an enhanced mask decoder method. In order to help model considering all replies under the same idx as the same importance, post-processing technique was given. Like majority voting, a voting technique was added to replace all labels to the major label under the same idx. The development dataset from Fake-EmoReact 2021 was applied as a validation dataset, which is for fine-tuning hyperparameters. The best tuning model predicted the results for the evaluation dataset. 

\paragraph{Jina}
Multiple machine learning models were used to detect fake news, including LR, decision tree, XGBoost, CNN, and BiLSTM. Analyzing the textual characteristics of Twitter posts and conducting the preprocessing procedures enhances the model performance. In round 2, they used LR model to detect fake news with a superior accuracy (99.05\%). They also revealed that machine learning models perform comparatively better than the other deep learning models with relatively lower computational complexities.

\paragraph{TeamZulu}
Various deep learning models were implemented with multiple Hyper-parameters settings, including CNN, RNN-based model, BERT, and RoBERTa. They trained an ensemble framework to aggregate the
predictions of each model and results in a final prediction. A pre-trained GLoVe and three voting strategies were also added to improve model performance. A pre-trained Bi-LSTM model predicted final results for round two. An ablation study and model comparison was also discussed in their study.

\section{Evaluation \& Discussion}
\begin{table*}[t]
\centering
\begin{tabular}{@{}lllcrrrr@{}}
\toprule
\textbf{Rank} & \textbf{Team} & \textbf{F1-macro} & \textbf{Precision} & \textbf{Recall}  \\ 
\midrule
1 & Yao & 93.09 & 93.59 & 92.88\\
2 & Jina & 82.75 & 83.43 & 82.57 \\
3 & SpencerChen & 76.83 & 78.56 & 77.67 \\
4 & TeamZulu & 74.99 & 76.39 & 45.74 \\
5 & skblaz & 40.77 & 73.39 & 54.24 \\

\bottomrule
\end{tabular}
\caption{F-scores for round two evaluation dataset without (Other, None) categories (\%)}
\label{tab:without_empty}

\end{table*} 
A summary of submissions and the final results, submitted by participants in the Fake-EmoReact challenge, are shown in Table \ref{tab:model}.
Macro-F1 score from all five teams was higher than 70\%. Pre-trained models and TF-IDF was used by the top teams, and deep learning models received outstanding performances from Fake-EmoReact 2021 dataset. 
The top score for the share task is 93.9\%, 93.46\%, 94.74\% for F1-score, Precision, and Recall, respectively.
Full Leaderboard from both rounds is available on Fake-EmoReact 2021 website. More observations from the challenge and submissions was revealed below. 

\paragraph{Deep learning model}
All of the team tries a least one of the deep learning models in their submission. Two of the team applies a model relative to BERT, with pre-trained embedding. Some of the participants enhanced the model's performance by adding voting strategies and textual characteristics. We observe well-performed models require excellent
exploratory data analysis, and domain knowledge in the share task.

\paragraph{Imbalanced categories}
Fake-EmoReact 2021 dataset has suffered categories imbalance situation since the negative samples were selected from EmotionGIF 2020 dataset \cite{shmueli2021socialnlp}. Most of the negative samples contain more than one GIF category in a source data. On the other hand, only two categories (Other, None) are provided in a positive sample. Table \ref{tab:without_empty} shows the model performance in round two, without data containing only two categories (Other, None), from all teams. A performance decrease happened in user skblaz's model; others remain similar results.
We plan to add more samples containing only other or none categories in the negative sample to solve this problem.

\section{Conclusion}

We successfully manage a fake new detection challenge, Fake-EmoReact 2021 , in SocialNLP 2021. Many researchers have noticed this challenge and requested the datasets. Furthermore, 5 submitted their results successfully in the evaluation phase. Various deep learning and machine learning model are applying in the share task. The best team achieve 93.9 \% macro-F1 score in final round.

\section{Acknowledgment}
This research is partially supported by Ministry of Science and Technology, Taiwan, under Grant no. MOST109-2221-E-001-015- and MOST108-2221-E-001-012-MY3.

\bibliography{anthology,custom}
\bibliographystyle{acl_natbib}


\end{document}